# Multi-Scale Label Relation Learning for Multi-Label Classification Using 1-Dimensional Convolutional Neural Networks

Junhyung Kim[†]
School of Computer Science and
Electrical Engineering,
Handong Global University
Pohang, South Korea
21800180@handong.edu

Byungyoon Park
School of Computer Science and
Electrical Engineering,
Handong Global University
Pohang, South Korea
21800271@handong.edu

Charmgil Hong
School of Computer Science and
Electrical Engineering,
Handong Global University
Pohang, South Korea
charmgil@handong.edu

## ABSTRACT

We present Multi-Scale Label Dependence Relation Networks (MSDN), a novel approach to multi-label classification (MLC) using 1-dimensional convolution kernels to learn label dependencies at multi-scale. Modern multi-label classifiers have been adopting recurrent neural networks (RNNs) as a memory structure to capture and exploit label dependency relations. The RNN-based MLC models however tend to introduce a very large number of parameters that may cause under-/over-fitting problems. The proposed method uses the 1-dimensional convolutional neural network (1D-CNN) to serve the same purpose in a more efficient manner. By training a model with multiple kernel sizes, the method is able to learn the dependency relations among labels at multiple scales, while it uses a drastically smaller number of parameters. With public benchmark datasets, we demonstrate that our model can achieve better accuracies with much smaller number of model parameters compared to RNN-based MLC models.

## 1 Introduction

In machine learning, classification is to learn the relationship between a set of observations and its associated output variable (a label or class). The conventional assumption for classification is that each data instance is associated with a single label variable. In many real-world data analyses, however, there are cases where each data instance is associated with two or more labels. For example, when one classifies movie genres, a movie can fall into the categories of *comedy* and *adventure* at the same time. Likewise, a news article that covers a celebrity throwing out ceremonial first pitch can be classified as *sports* and *celebrity* news article. As in the above cases, we often need more generalized notions and definitions than the conventional classification setting to properly handle certain real-world problems.

*Multi-label classification*, or MLC in short, is a classification problem where each data instance can be associated with multiple label variables. Accordingly, the goal of MLC is to produce a predictive model that can accurately map the input observations to the correct sets of labels. That is, MLC aims at correctly predicting all the labels assigned to a data instance. One way to approach this problem is to learn a model that can represent the joint probability distribution of the label space given observations.

One of the most intuitive solutions to MLC is to decompose an MLC problem into multiple binary classification problems [1]. Namely, for a data instance, each binary classifier predicts whether the data belongs to a specific label or not. However, this approach holds a strong assumption that each label is conditionally independent with each other, that usually does not hold in real-world problems. Another intuitive solution is to transform an MLC problem into a multi-class classification problem [2]. That is, each class in the transformed problem represents one possible combination of label sets. While this solution can be advantageous in that the dependence relations among labels can be fully considered, it requires an exponential space complexity that makes the solution less preferable.

To take both the label dependence relations and cost into account while addressing the MLC problem, more structured approaches have been introduced. According to the model properties, these approaches can be categorized into two groups: the *chaining* and *stacking* methods. The chaining methods [3,4,5] are to infer the MAP (maximum a posteriori) assignment of the label variables via a probabilistic decomposition using the chain rule of probability. On the other hand, the stacking methods [6] build multiple levels of classifiers to indirectly take the interactions between labels into account. Recently, deep learning approaches have been proposed to represent non-linear relationships among input and label variables. More specifically, as the memory structure of recurrent neural networks (RNN) is considered to be useful in capturing and exploiting the dependence relations among labels, several RNN-based models [7,8,9] were introduced. However, these models tend to produce a very large number of model parameters that are often subject to under-/over-fitting problems.

In this paper, we propose a novel deep neural network-based approach for MLC, which can effectively learn conditional dependence relations between labels. In particular, the proposed model adopts 1-dimensional convolutional neural networks (1D-CNN) with multiple kernels to learn the label dependence relations, while yielding a significantly smaller number of model parameters, compared to the existing RNN-based MLC models. By setting 1D-CNN kernels with different kernel sizes, our model can capture the label dependence relations in the multiple scales.

To our best of knowledge, the proposed solution includes the first model that utilizes the 1D-CNN structure to learn and utilize the label dependences for tackling the MLC problem. In Section 2, we review the background required for our discussion. Section 3 present the proposed method. Section 4 reports the experimental results and analyses. Section 5 concludes the paper.

## 2 Background

### 2.1 Problem Definition

*Multi-Label Classification* is a classification problem where each data instance is associated with a set of labels. This paper denotes the number of instances, the number of features, and the number of labels by $N, m,$ and $d$, respectively. Given a dataset $D = \{\mathbf{x}^{(n)}, \mathbf{y}^{(n)}\}_{n=1}^{N}$, where $\mathbf{x}^{(n)} = \left(x_1^{(n)}, x_2^{(n)}, \dots, x_m^{(n)}\right)$ is the $n$-th data instance with $m$ features, and $\mathbf{y}^{(n)} \in \{0,1\}^d$ is a $d$-dimensional binary vector representing a set of labels associated with the $n$-th data instance. The goal of MLC is to learn a classifier $h$ from $D$, which takes $\mathbf{X}$ as input and produces prediction of $\mathbf{Y}, \widehat{\mathbf{Y}}$; that is, $h: \mathbf{X} \rightarrow h(\mathbf{X}) = \widehat{\mathbf{Y}}$.

### 2.2 Previous Research

This section briefly reviews the existing MLC studies that are related to our work. One of the simplest approaches to MLC is Binary Relevance (BR) [1] that transforms an MLC problem into multiple binary classification problems. Accordingly, by design, each BR optimizes the marginal probability of each class label. In addition to that, BR takes a strong assumption that the labels are independent to each other, which does not hold in most MLC cases. As the goal of MLC is to learn a classifier that optimizes the joint probability distribution of the entire label space, a key to tackle the problem is to capture the label dependence relations from data. To this end, Label Powerset (LP) [2] has been introduced. That is, LP transforms an MLC problem to a multi-class classification problem, each class in LP represents a possible combination of labels. While the method is straightforward, LP is known to be impractical due to the exponential space complexity over the number of labels.

Later, more structured approaches have been proposed to sort out the MLC problem by modeling the label dependences in systematic manners. As introduced in Section 1, these methods can be divided into the *chaining* and *stacking* methods. Classifier Chain (CC) [3] is one of the representatives of the former group that decomposes a joint probability distribution of the labels given input observations, with the chain rule of probability. As a result, CC effectively optimizes the conditional joint distribution of the label space, while its procedure and cost are very similar to BR. Probabilistic Classifier Chain (PCC) [4] further optimizes the method by tackling the point that the accuracy of CC can be significantly influenced by the label order that the model follows. While PCC is capable to yield the optimum results, as it adopts an exhaustive search for prediction that requires $O(2^d)$ time, the method is not considered to be practical.

On the other hand, an example of the stacking methods is Stacked BR (STA) [6]. Unlike the chaining methods, STA seeks a fully-connected relationship between labels, by stacking two levels of the BR-style MLC models. At the first level, it learns the relationship between an observation and its associated label variables. Then, at the second level, it takes the raw outputs of first level as additional features to exploit label relationships.

Recent years have seen increased interests in the neural network-based MLC approaches. These methods are intended to learn higher-order features that represent the relationships among input and label variables. The recurrent neural network (RNN) structure is particularly getting attention as it has shown its capability to capture the label dependence relations in the MLC problems. That is, Read and Jaakko [10] claimed that CC can be seen as a reinterpreting process of carrying previous prediction variables, and reimplemented CC with RNN as memory units. CNN-RNN [7] is another method that adopts RNN to represent the chaining structure. By combining the CNN and RNN structures, the authors facilitated a model that learns the relationships among the label variables using RNN, whereas the CNN structure is used to learn feature representations from image data. Later, Att-RNN [8] has been proposed to capture the label dependences more effectively with the attention mechanism [11]. While the above models utilize the recurrent architecture of RNN to capture the dependence relations in the label space, RethinkNet [9] presented a different approach to adopt RNN. The proposed "rethinking" structure implements iteratively improves the quality of multi-label prediction.

One remark on the RNN-based models is that those models tend to introduce a very large number of model parameters. This tendency may easily lead the model to under-/over-fitting problem. In the next section, we present our model that uses 1-Dimensional Convolution Neural Network (1D-CNN) to learn the label relationships. By using 1D-CNN kernels, our model can effectively learn the label dependence relations from data while it produces much smaller number of parameters than RNN-based models.

## 3 Methods

### 3.1 1D Convolutional Neural Network (1D-CNN)

1D-CNN [12] is able to extract the features required for prediction by pooling the values obtained through the convolution operation. A kernel in the 1D-CNN layer performs a convolution function between input and parameters while sliding the kernel from the left end of the input vector to the other end. The results of convolution between input and kernel are passed through the activation function and only a subset of the outputs is selected after the pooling layer, that is to generate a new feature representation for classification.

**Figure 1: Multi-Scale Label Dependence Relation Networks**

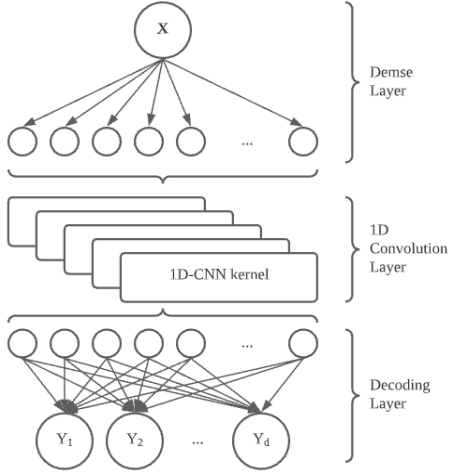

**Table 1: Dataset characteristics**

| Datasets | #Instances | #Features | #Labels | Domain |
|---|---|---|---|---|
| Scene | 2408 | 294 | 6 | image |
| Yeast | 2417 | 103 | 14 | biology |
| Business | 11214 | 21950 | 30 | text |
| Science | 6428 | 37230 | 40 | text |
| TMC2007_500 | 28596 | 500 | 22 | text |

When there is no zero padding in input data, the size of feature map is $c = \frac{mk}{s} + 1$, which is determined by input size $m$, kernel size $k$, and kernel stride $s$. Each element of the feature map can be written as $o_j = \sigma(\sum_{i=1}^{k} w_i x_{i+(j-1)s})$ for $1 \leq j \leq c$. Here, σ denotes an activation function, and the sigmoid function is used in our experiments.

## 3.2 Multi-Scale Label Dependence Relation Networks

We propose Multi-Scale Label Dependence Relation Networks (MSDN) which is a new MLC model with 1D-CNNs. Specifically, MSDN inherits the base structure from the stacking methods [6] while it connects the two layers of the model components by adopting 1D-CNN kernels in between the layers, and extracts the vectors representing dependence relations among labels. Figure 1 illustrates the architecture of MSDN. The model consists of three parts: *Dense* (fully-connected) layer for creating a mid-level representation of label vector, *1-D Convolution* layer for capturing the dependencies between labels, and *Decoding* layer for decoding outputs from *1-D Convolution* layer to the label vectors.

The purpose of the *Dense* layer is to learn the mappings between a data instance and a mid-level representation of the label vector. That is, through the training process, the output of the *Dense* layer learns an intermediate representation between the observation and its associated labels. The *1-D Convolution* layer takes the mid-level representation from the *Dense* layer as an input. Using multiple 1-dimensional CNN kernels with different sizes, the *1-D Convolution* layer can extract the embedding of dependence relations among neighboring labels. That is, each kernel seeks a potential relation among different number of labels. Finally, we use the max-pooling operation to pick up the best signal. Accordingly, our model construction enables learning of various dependence relation information at different scales by setting the size of each convolution kernel differently. Lastly, the *Decoding* layer learns the mapping between the embedded vector and the label vector, which can be interpreted as a decoding process from the output of the *1-D Convolution* layer to the original label vector.

## 4 Experiment

To validate the effectiveness of our proposed model, we conduct experiments using 5 public datasets [1,13,14,15]. We split each dataset into 75% of training and 25% of testing, randomly for 5 times. We apply the min-max scaling to put all the feature values ranging between 0 and 1. We compare the performance of MSDN with several MLC baselines: Binary Relevance (BR) [1], Classifier Chain (CC) [3], Probabilistic Classifier Chain (PCC) [4], Stacked BR (STA) [6], CC-RNN, and RethinkNet [8]. Note that CC-RNN is a variation of CNN-RNN [7], which excludes the CNN layer from the original CNN-RNN structure [8]. To evaluate the model performances, we use EMA (exact match accuracy), which computes the percentage of exact matches between predicted vectors and true label vectors; and micro-average F1 Score, the harmonic mean of precision and recall after summing up all the individual true positive, false positive, and false negative, as the key metrics. We compare the results using the paired t-test at $\alpha = 0.05$, and report which models are winning and losing.

In our experiments, the size of all the hidden vectors is fixed to 128. Our model also uses 128 1D-CNN kernels (where the size of $i$-th kernel is $i$) with max-pooling layers to form up *1-D Convolution* layer. We used the Adam optimizer with the binary cross-entropy loss function. We selected hyperparameters with the best performance in terms of EMA. The search space considered includes – learning rate: {0.0005, 0.00075, 0.001, 0.0025, 0.005, 0.0075, 0.01, 0.025, 0.05, 0.075}, drop out: {0, 0.25, 0.5}, weight decay: {0, 0.00001, 0.000025, 0.00005, 0.000075, 0.0001}. The batch size was fixed to 128. We trained the model with the early stopping technique, where the maximum epoch is set to 10,000.

Table 2 and 3 show the EMA and micro-average F1 score of our benchmarking models respectively. The results of PCC are not available on Business, Science, and TMC2007_500 since their experiment time exceeded 24 hours for a round. On both EMA and micro-average F1 score, we can see that our model outperformed other competitors for most datasets. MSDN won all the competitors from Business and TMC2007_500 datasets for both EMA and micro-average F1 score. In terms of EMA, our model significantly outperformed BR and CC for all datasets, significantly outperformed STA and CC-RNN for 4 datasets. Likewise, in terms of micro-average F1 score, our model significantly better than BR and CC-RNN for all the datasets, significantly better than CC for 4 datasets, significantly better than CC for 3 datasets. The exceptions are RethinkNet and PCC, where the number of ties is greater or equal to the number of wins for both EMA and micro-average F1 score. Although they showed similar performance to MSDN, note that they never showed better performance than MSDN. In addition, the performance of RethinkNet on Science dataset was very unstable, the performances recorded for each round were very different.

Table 4 shows the number of parameters introduced from each MSDN and other deep learning MLC methods. On average, the number of parameters from MSDN is 124,808 less than the number of parameters from RethinkNet, 525,449 less than the number of parameters from CC-RNN. These results state that our model outperformed other deep learning-based MLC models with much fewer parameters. This indicates that our model can learn label dependence relations more efficiently than other deep learning-based MLC methods.

## 5 Conclusion

In this paper, we proposed a novel deep learning-based method for multi-label classification. Unlike other deep learning-based MLC models, our model adopted 1 Dimensional Convolution Neural Network (1D-CNN) to learn the label dependence relations from data. Our model conducts modeling label dependence relationships in various scales using multiple 1D-CNN kernels with different kernel sizes for each kernel. Furthermore, from benchmarking with other well-known MLC methods, our model outperformed other competitors in most cases in terms of EMA and micro-average F1 score. Also, from comparing the number of parameters and performances between MSDN and other deep learning-based models, MSDN proved that it could learn label dependence relations efficiently with much fewer parameters than other comparable models.

Table 2: Experiment results of each method in terms of EMA(exact match accuracy). Symbols ⊖/† indicates whether MSDN statistically better/worse to the competitor (using paired t-test at 0.05 significance level). The last row/column indicates win/tie/loss for MSDN on each datasets/against the compared model.

| Models | Scene | Yeast | Business | Science | TMC2007_500 | #win/#tie/#loss |
|---|---|---|---|---|---|---|
| BR | 0.5256±0.0133 ⊖ | 0.1332±0.0133 ⊖ | 0.5459±0.0122 ⊖ | 0.2668±0.0077 ⊖ | 0.3110±0.0040 ⊖ | 5 / 0 / 0 |
| CC | 0.6236±0.0133 ⊖ | 0.1521±0.0138 ⊖ | 0.5513±0.0053 ⊖ | 0.2978±0.0091 ⊖ | 0.3259±0.0032 ⊖ | 5 / 0 / 0 |
| PCC | 0.6551±0.0109 ⊖ | 0.1934±0.0234 | - | - | - | 1 / 1 / 0 |
| STA | 0.6156±0.0121 ⊖ | 0.1620±0.0135 ⊖ | 0.5516±0.0073 ⊖ | 0.3323±0.0130 | 0.3181±0.0021 ⊖ | 4 / 1 / 0 |
| CC-RNN | 0.6233±0.0179 ⊖ | 0.1736±0.0116 | 0.5586±0.0126 ⊖ | 0.2111±0.0101 ⊖ | 0.3087±0.0073 ⊖ | 4 / 1 / 0 |
| RethinkNet | 0.6844±0.0111 | 0.1904±0.0119 | 0.5462±0.0075 ⊖ | 0.2180±0.1288 | 0.3106±0.0025 ⊖ | 2 / 3 / 0 |
| MSDN | 0.6764±0.0147 | 0.1881±0.0157 | 0.5730±0.0072 | 0.3323±0.0101 | 0.3585±0.0059 | - |
| #win/#tie/#loss | 5 / 1 / 0 | 3 / 3 / 0 | 5 / 0 / 0 | 3 / 2 / 0 | 5 / 0 / 0 | |

Table 3: Experiment results of each method in terms of micro-average F1 score. Symbols ⊖/† indicates whether MSDN statistically better/worse to the competitor (using paired t-test at 0.05 significance level). The last row/column indicates win/tie/loss for MSDN on each datasets/against the compared model.

| Models | Scene | Yeast | Business | Science | TMC2007_500 | #win/#tie/#loss |
|---|---|---|---|---|---|---|
| BR | 0.6948±0.0120 ⊖ | 0.6331±0.0091 ⊖ | 0.6835±0.0176 ⊖ | 0.3339±0.0213 ⊖ | 0.6987±0.0046 ⊖ | 5 / 0 / 0 |
| CC | 0.7218±0.0132 ⊖ | 0.6143±0.0139 ⊖ | 0.6861±0.0036 ⊖ | 0.3807±0.0109 | 0.6980±0.0041 ⊖ | 4 / 1 / 0 |
| PCC | 0.7355±0.0166 ⊖ | 0.6464±0.0113 | - | - | - | 1 / 1 / 0 |
| STA | 0.7309±0.0118 ⊖ | 0.6378±0.0083 | 0.6908±0.0052 ⊖ | 0.4258±0.0134 | 0.7034±0.0050 ⊖ | 3 / 2 / 0 |
| CC-RNN | 0.6824±0.0120 ⊖ | 0.6212±0.0146 ⊖ | 0.6987±0.0100 ⊖ | 0.2449±0.0215 ⊖ | 0.6800±0.0160 ⊖ | 5 / 0 / 0 |
| RethinkNet | 0.7529±0.0113 | 0.6451±0.0072 | 0.6838±0.0052 ⊖ | 0.2776±0.2079 | 0.6889±0.0071 ⊖ | 2 / 3 / 0 |
| MSDN | 0.7516±0.0131 | 0.6483±0.0095 | 0.7263±0.0055 | 0.4090±0.0193 | 0.7251±0.0065 | - |
| #win/#tie/#loss | 5 / 1 / 0 | 3 / 3 / 0 | 5 / 0 / 0 | 2 / 3 / 0 | 5 / 0 / 0 | |

Table 4: Number of parameters introduced from each models for each datasets

| Models | Scene | Yeast | Business1 | Science1 | TMC2007_500 |
|---|---|---|---|---|---|
| CC-RNN | 566,919 | 545,551 | 3,344,799 | 5,302,313 | 599,447 |
| RethinkNet | 170,630 | 100,296 | 2,942,366 | 4,897,320 | 199,062 |
| MSDN | 23,502 | 46,918 | 2,816,719 | 4,770,383 | 524,097 |